\pdfoutput=1

\documentclass[11pt]{article}

\usepackage{acl}

\usepackage{times}
\usepackage{latexsym}

\usepackage[T1]{fontenc}

\usepackage[utf8]{inputenc}

\usepackage{microtype}
\usepackage{xcolor} 
\usepackage{todonotes}
\usepackage{multirow}
\usepackage{graphicx}
\usepackage{rotating}
\usepackage{todonotes}
\usepackage{amsmath,amssymb}
\usepackage{siunitx}
\usepackage{booktabs}
\usepackage{comment}
\usepackage{pifont}
\newcommand{\cmark}{\ding{51}}
\newcommand{\xmark}{\ding{55}}

\newcommand{\word}[1] {`\textit{#1}'}
\newcommand{\sense}[1] {`\textsc{{#1}}'}

\newcommand{\hide}[1]{}

\title{Do Not Fire the Linguist:\\Grammatical Profiles Help Language Models Detect Semantic Change}

\author{Mario Giulianelli\thanks{~~Equal contribution, the authors are listed alphabetically.} \\
  ILLC,
  University of Amsterdam \\
  \texttt{m.giulianelli@uva.nl} \\\And
  Andrey Kutuzov\footnotemark[1] \\
  University of Oslo \\
  \texttt{andreku@ifi.uio.no}\\\And
  Lidia Pivovarova \\
  University of Helsinki \\
  \texttt{first.last@helsinki.fi}
  }

\begin{document}
\maketitle

\begin{abstract}
Morphological and syntactic changes in word usage---as captured, e.g., by grammatical profiles---have been shown to be good predictors of a word's meaning change.
In this work, we explore whether large pre-trained contextualised language models, a common tool for lexical semantic change detection, are sensitive 
to such morphosyntactic changes. To this end, we first compare the performance of grammatical profiles against that of a multilingual neural language model (XLM-R) on 10 datasets, covering 7 languages, and then combine the two approaches in ensembles to assess their complementarity. Our results show that 
ensembling grammatical profiles with XLM-R improves semantic change detection performance for most datasets and languages. 
This indicates that language models do not fully cover the fine-grained morphological and syntactic signals that are explicitly represented in grammatical profiles.

An interesting exception are the test sets where the time spans under analysis are much longer than the time gap between them (for example, century-long spans with a one-year gap between them). 
Morphosyntactic change is slow so grammatical profiles do not detect in such cases. In contrast, language models,  thanks to their access to lexical information, are able to detect fast topical changes.
\end{abstract}


\section{Introduction}
\label{sec:introduction}

Human language is in continuous evolution. New word senses arise, and existing senses can change or disappear over time as a result of social and cultural dynamics or technological advances. NLP practitioners have become increasingly interested in this diachronic perspective of semantics. Some works focus on constructing, testing and improving psycholinguistic and sociolinguistic theories of meaning change \cite{xu2015computational,hamilton-etal-2016-diachronic,goel2016social,noble-etal-2021-semantic}; 
others are concerned with surveying how the meaning of words has evolved historically \cite{garg2018word,kozlowski2019geometry} or how it is currently transforming in public discourse \cite{azarbonyad2017words,del-tredici-etal-2019-short}. Recently, we also see increased interest in more application-oriented work, with efforts to develop adaptive learning systems that can remain up-to-date with humans' continuously evolving language use~\cite[\textit{temporal generalization};][]{lazaridou2021mind}.

An increasingly popular way to determine whether and to what degree the meaning of words has changed over time is to use `contextualised' (or `token-based') word embeddings extracted from large pre-trained language models~\cite{giulianelli-etal-2020-analysing,montariol-etal-2021-scalable} as they encode rich, context-sensitive semantic information.
However, it has also been shown recently that changes in the frequency distribution of morphological and syntactic features of words, as captured by \textit{grammatical profiles}, can also be employed for lexical semantic change detection~\cite{giulianelli-etal-2021-grammatical}, with competitive performance.
These are, to some extent, two opposing approaches: while language models (LMs) are largely based on word co-occurrence statistics, grammatical profiles are de-lexicalised and rely on explicit linguistic information. 

Although they are superficially unaware of morphology and syntax, LMs have been shown to capture approximations of grammatical information in their deep representations~\cite{warstadt-etal-2020-blimp-benchmark}. Yet are these sufficient to detect meaning shifts that are accompanied by morphosyntactic changes in word usage?
We hypothesise that this is not the case, and to test this hypothesis, we combine LM-based methods and grammatical profiles into ensemble models of lexical semantic change detection.\footnote{Throughout the paper, we refer to the systems that combine LMs with grammatical profiles as `ensembles'. These are not statistical methods of \textit{ensemble learning} but systems that combine the predictions of different models.} If adding grammatical profiles to LMs results in a boost in performance, then this means that LMs do not capture morphosyntactic change as accurately as explicit morphological tagging and syntactic parsing (or at the very least that it is difficult to extract this type of information from the models). If we do not observe any boost, this suggests that LMs already represent all the necessary grammatical information and explicit linguistic annotation is not required. 
We conduct our experiments with 10 datasets, covering 7 languages. For comparability, we use the same model for all the languages. We choose XLM-R~\cite{conneau-etal-2020-unsupervised}, a multilingual Transformer-based masked language model which has already been successfully applied to the semantic change detection task~\cite{arefyev-zhikov-2020-bos,arefyevdeepmistake}. Although it covers the full linguistic diversity of our data, we additionally fine-tune XLM-R on monolingual diachronic corpora.

\begin{figure}
    \centering
    \includegraphics[width=\linewidth]{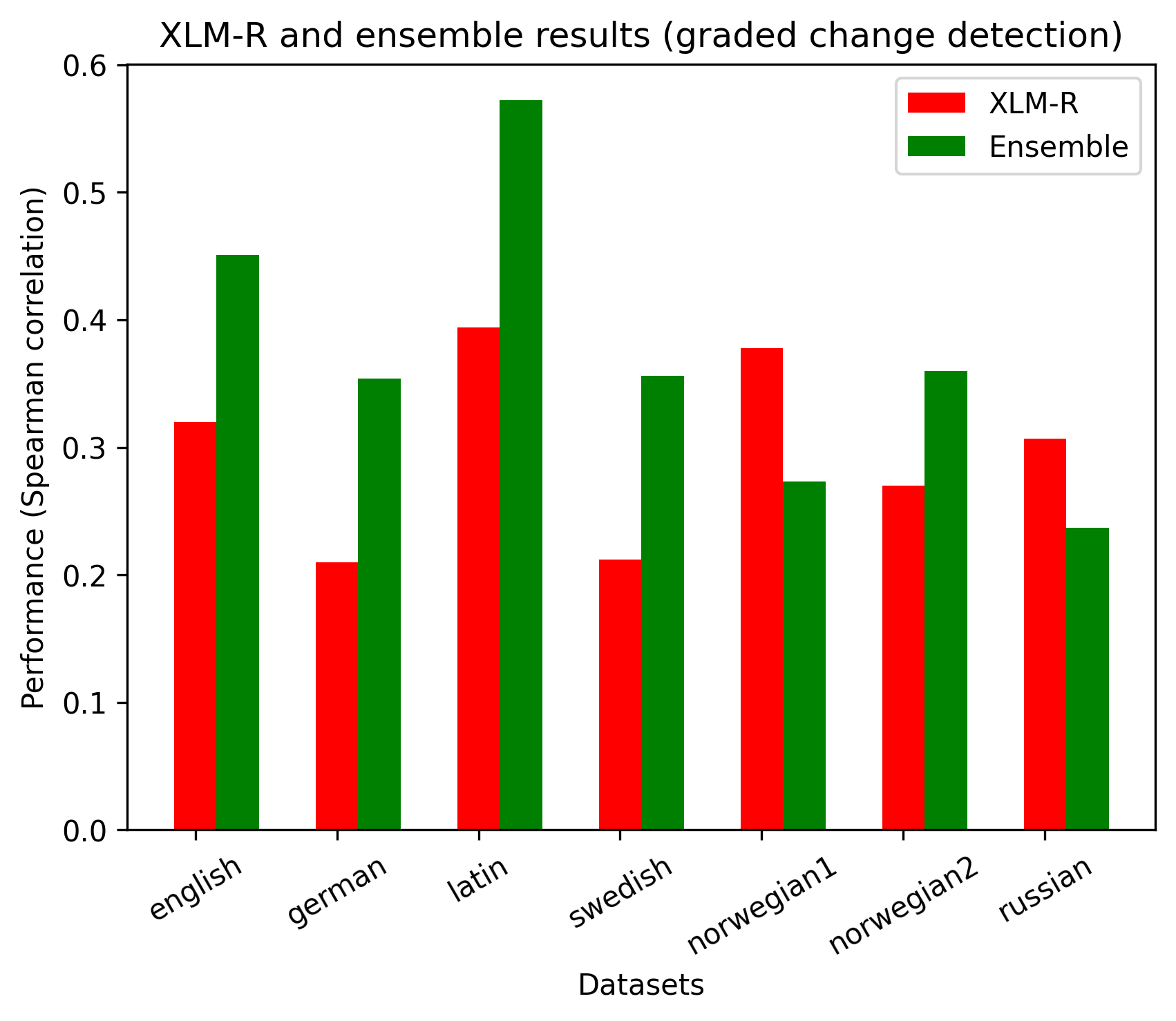}
    \caption{Performance of an XLM-R based method (PRT) and an ensemble method (PRT-MORPHSYNT) on the ranking task; see Section~\ref{sec:methods} for method descriptions. The scores for the three Russian datasets are averaged as they exhibit similar trends.}
    \label{fig:deltas}
\end{figure}

Our quantitative and qualitative evaluation of the resulting ensembles on the graded and binary semantic change detection tasks largely confirm our hypothesis. Ensembling XLM-R and grammatical profiles improves the results for 4 out of 6 languages in graded change detection (as well as for 1 of the 2 Norwegian datasets) and for 5 out of 6 languages in binary change detection. 
Figure~\ref{fig:deltas} illustrates these improvements.
The reasons why ensembles do not outperform the XLM-R baseline on some datasets are linked to the size of the gaps between the historical time periods represented in the diachronic corpora; we analyse and discuss these reasons in Section~\ref{sec:results-why}. Overall, we show that providing large language models with explicit morphosyntactic information helps them quantify semantic change.  


\section{Tasks and Data}
\label{sec:tasks}
The goal of lexical semantic change detection is to determine whether and to what extent a word's meaning has changed over a certain period of time. The performance of automatic systems that address this problem is typically assessed in two tasks~\cite{schlechtweg-etal-2020-semeval}.
\textbf{Task 1} is a binary classification task: given a diachronic corpus and a set of target words, a system must determine whether the words lost or gained any senses between two time periods. We refer to it as the \textit{classification task} and use accuracy as an evaluation metric. \textbf{Task 2} is a ranking task: a system must rank the target words according to the degree of their semantic change. We refer to it as the \textit{ranking task} and use the Spearman rank-correlation with the gold rankings as an evaluation metric.

We rely on a collection of diachronic corpora and annotated target word lists covering seven languages from three Indo-European language families. Target words are annotated with binary and graded scores of semantic change, corresponding respectively to Task 1 and 2~\cite{schlechtweg-etal-2018-diachronic}. English (EN), German (DE), Latin (LA), and Swedish (SW) data are available from the SemEval 2020 Unsupervised Lexical Semantic Change Detection shared task~\cite{schlechtweg-etal-2020-semeval}. For Italian (IT), we use the data released for the EvaLita competition~\cite{basile2020diacr}. For Norwegian (NO), we use the NorDiaChange dataset recently released by \citet{kutuzov2022nordiachange}, consisting of two subsets with different target word lists and time spans. Finally, for Russian (RU), we draw from the RuShiftEval shared task~\cite{rushifteval2021}, consisting of three subsets with different time spans and a shared target word list. 
Table~\ref{tab:data} summarises the most important properties of the datasets and indicates what types of annotations are available for each language. Note that the subset splitting in Norwegian and Russian is not introduced by us, but is provided by the corresponding dataset creators.

\begin{table*}
    \centering
    \resizebox{\textwidth}{!}{
    \begin{tabular}{l|cccccccccc}
        & \textbf{EN}	& \textbf{DE} & \textbf{IT}	& \textbf{LA}	& \textbf{NO-1}	& \textbf{NO-2}	& \textbf{RU-1}	& \textbf{RU-2} & \textbf{RU-3} & \textbf{SW} \\
        \hline
        \textbf{Period 1}        & 1810-1860 & 1800-1899 & 1945-1970 & -200-0 & 1929-1965 & 1980-1990 & 1700-1916 & 1918-1990 & 1700-1916 & 1790-1830 \\
        \textbf{Period 2}        & 1960-2010 & 1946-1990 & 1990-2014 & 0-2000 & 1970-2013 & 2012-2019 & 1918-1990 & 1992-2016 & 1992-2016 & 1895-1903 \\
        \textbf{Tokens} (mln)         & 7+7 & 70+72 & 52+197 & 2+9 & 57+175 & 43+649 & 93+122 & 122+107 & 93+107 & 71+110 \\
        \textbf{Targets}         & 37 & 48 & 18 & 40 & 80 & 80 & 99 & 99 & 99 & 32 \\
        \textbf{Ranking}         & \cmark & \cmark & \xmark & \cmark & \cmark & \cmark & \cmark & \cmark & \cmark & \cmark \\
        \textbf{Classification}  & \cmark & \cmark & \cmark & \cmark & \cmark & \xmark & \xmark & \xmark & \xmark & \cmark \\
    \end{tabular}}
    \caption{Statistics for our collection of diachronic corpora and of the corresponding semantic change annotations.}
    \label{tab:data}
\end{table*}

\section{Methods}
\label{sec:methods}

\subsection{Grammatical Profiles}
\label{sec:method-profiles}

Grammatical profiling is a corpus linguistic technique which allows to distinguish subtle semantic differences by measuring the distance between distributions of grammatical parameters~\cite{gries2009behavioral,janda2011grammatical}. It has been shown recently that diachronic changes in grammatical profiles can serve as a strong indication of semantic change~\cite{giulianelli-etal-2021-grammatical}. For our experiments we adopt this method in its best performing configuration.

First, the diachronic corpus of interest is tagged and parsed using UDPipe~\cite{straka-strakova-2017-tokenizing}. We then find all occurrences of a target word in the corpus and create a count vector for each detected morphological feature. 
For example, a morphological profile for an English verb could look as follows: \vspace{0.3em}\\
\texttt{\small
\hspace*{0.2cm}Tense : \{Past 42, Pres 51\}\\
\hspace*{0.2cm}VerbForm : \{Part 68, Fin 25, Inf 9\}\\
\hspace*{0.2cm}Mood : \{Ind 25\}\\
\hspace*{0.2cm}Voice : \{Pass : 17\}\vspace{0.4em}\\
}
In this way, count vectors are constructed for each target word in each time period of the corpus; these are a word's grammatical profiles. 
The cosine distance between count vectors is computed separately for every morphological category, and the degree of semantic change between periods is measured as the maximum among the computed cosine distances. We refer to this type of grammatical profile as \textbf{MORPH}. 

In addition to morphological features, a separate vector of syntactic features is created, which contains counts of dependency arc labels from a target word to its syntactic head. We refer to these grammatical profiles as \textbf{SYNT}. Semantic change is measured as the cosine distance between two syntactic vectors.
%
Morphological and syntactic profiles can also be combined. We do this by concatenating syntactic features to the array of morphological features, and then using the maximum cosine distance as our third profile-based measure of semantic change, \textbf{MORPHSYNT}.

\subsection{Static Embeddings}
\label{sec:method-static}
Static embeddings~\cite[e.g.,][]{Mikolov_representation:2013} are known to perform very well at detecting lexical semantic change~\cite{schlechtweg-etal-2020-semeval}. Therefore, although they are not directly relevant to our research question, we include them in our experiments as a point of comparison, following the common approach proposed by~\citet{hamilton-etal-2016-diachronic}. Further details can be found in Appendix~\ref{sec:app-static}.

\subsection{Contextualised Embeddings}
\label{sec:method-contextualised}
Many have argued that static representations are not theoretically appropriate as a model of word meaning because they conflate all the usages of a word into a single context-independent embedding, and that contextualised representations should be used instead~\cite[e.g.,][]{schutze-1998-automatic,erk-pado-2008-structured,pilehvar-collier-2016-de}. This has motivated the development of semantic change detection algorithms that rely on context-dependent representations, where every usage of a word corresponds to a unique \textit{token embedding}~\cite{giulianelli-etal-2020-analysing,martinc-etal-2020-leveraging}.
Language models produce very competitive results across languages~\cite{kutuzov-giulianelli-2020-uio}, and they lead to more interpretable systems~\cite{montariol-etal-2021-scalable}. 

We choose XLM-R~\cite{conneau-etal-2020-unsupervised} as our pre-trained language model, since it was shown to perform well in semantic change shared tasks~\cite{schlechtweg-etal-2020-semeval,rushifteval2021} and because, being multilingual, it can be applied to all languages under analysis, making evaluation more consistent.
First, we finetune XLM-R on the monolingual diachronic corpora of interest. Then, we deploy it to produce token embeddings for the target words in the diachronic corpus (in both time periods, T1 and T2). Further details on these two steps can be found in Appendix~\ref{sec:app-contextualised}.
%
We compute graded semantic change scores based on the extracted XLM-R embeddings and we use the scores to compile an ordered list of target words for the ranking task. Change scores are computed in four ways: 1) measuring the average pairwise cosine distance (\textbf{APD}) between embeddings collected in T1 and those in T2 \cite{giulianelli-etal-2020-analysing}; 2) measuring the cosine distance between prototype embeddings (\textbf{PRT})---i.e., the average contextualised word embeddings of T1 and T2 \cite{kutuzov-giulianelli-2020-uio}; 3) clustering the embeddings and then calculating the Jensen-Shannon Divergence (\textbf{JSD}) between the putative sense distributions of T1 and T2~\cite{martinc2020capturing,giulianelli-etal-2020-analysing}; 4) by taking a simple average (\textbf{APD-PRT}) of the predictions made by PRT and APD. The mathematical definitions of the metrics are given in Appendix~\ref{sec:app-metrics}.

\subsection{Change Point Detection}
\label{sec:method-changepoint}
To solve the classification task, we transform the continuous scores produced by our three metrics into binary semantic change predictions. Following \citet{giulianelli-etal-2021-grammatical}, we rank target words according to their continuous scores and classify the top $n$ words in the ranking as `changed' ($1$) and the rest of the list as 'stable' ($0$). To determine the change point $n$, we apply an offline change point detection algorithm ~\cite{truong2020selective} with the default settings.\footnote{\url{https://pypi.org/project/ruptures/}}

\subsection{Ensembling}
\label{sec:method-ensembling}
To find out whether grammatical profiles can improve the performance of embedding-based detection methods, we test all possible combinations of grammatical profile types and embedding-based metrics. Grammatical profiles come in three variants: MORPH, SYNT, and MORPHSYNT. Our embedding-based measures include APD, PRT, APD-PRT, and JSD. We compute the geometric mean $\sqrt{c_g c_e}$ between the change score $c_g$ obtained using grammatical profiles and the score $c_e$ output by an embedding-based metric, and use the resulting value as the ensemble semantic change score (e.g., \textbf{PRT-MORPHSYNT}).


\section{Results}
\label{sec:results}
We assess the performance of all methods presented in Section~\ref{sec:methods} on both semantic change detection tasks using our multilingual collection of semantic change datasets (see Section~\ref{sec:tasks}). 


\subsection{Ranking Task}
\label{sec:results-ranking}
For all methods, the Spearman rank-correlation between predicted scores and human annotations varies across languages and test sets; no method is a silver bullet for the ranking task (see Table~\ref{tab:ranking}).
XLM-R obtains higher correlation scores in English, Swedish, Norwegian-1, and Russian (1, 2, and 3); whereas grammatical profiles outperform it in German, Latin, Swedish, and Norwegian-2.
To better understand the strengths of all methods\hide{under scrutiny}, we now first present the results of each of them individually; then we report the performance of ensembles, where each method is combined with every other to generate semantic change predictions.

\paragraph{Grammatical Profiles}
Whether morphological features, syntactic features, or a combination of both are the most effective depends on the dataset; this also varies across test sets of the same language, as can be seen in the first three rows of Table~\ref{tab:ranking}. The performance of the different features diverges mostly for English and Norwegian-1, where SYNT is the best approach, as well as for Norwegian-2 and Russian, where MORPH works best. Combining morphological and syntactic features helps creating better rankings for German,
Latin, and Swedish. 

\paragraph{Contextualised Embeddings}
The correlation scores of average pairwise distance (APD) and prototype distance (PRT) differ substantially for all datasets, with the exception of Norwegian-1 (see rows 4-7 of Table~\ref{tab:ranking}). APD outperforms PRT on English, Swedish, Norwegian, and Russian; PRT is better on German and Latin.
Combining the two metrics in an ensemble (APD-PRT) marginally improves correlation scores for Norwegian-1 and Russian-1. Clustering contextualised embeddings (JSD) yields unstable results across datasets; it is the best contextualised method only for German.
    
\paragraph{Ensembles}
Whenever grammatical profiles produce better rankings than XLM-R, i.e., for German, Latin, Swedish, and Norwegian-2, combining the predictions of the two methods yields higher correlation scores than either method in isolation.
The most effective contextualised method in combination with grammatical profiles is PRT, regardless of the profile type. 
The PRT-MORPHSYNT combination produces the overall best ranking for German and Latin,
two languages with rich syntax and morphology. Which type of grammatical profile is the most complementary to XLM-R varies across datasets and it mostly corresponds to the profile type that obtains the best performance in isolation. Ensembles with JSD are outperformed by other methods, so we do not report them in Tables~\ref{tab:ranking} and \ref{tab:classification}.

%
Static embeddings, despite their good performance in isolation, do not combine well with grammatical profiles: this type of ensemble improves correlation scores only for Latin (Table~\ref{tab:ranking_sgns}).
As a final note, ensembles of grammatical profiles and XLM-R achieve the new best performance on the Latin ranking task of SemEval 2020 (PRT-MORPHSYNT), and establish a new SOTA for the recently released Norwegian-2 dataset (APD-MORPH).

\subsection{Classification Task}
\label{sec:results-classification}
Although our binary predictions for the classification task are dependent on the rankings discussed in the previous section, the overall trends of classification accuracy partly differ from the correlation trends of the ranking task. The classification results are shown in Table~\ref{tab:classification}. Compared to the ranking task, ensemble methods more often produce a performance improvement with respect to grammatical profiles and contextualised embeddings used in isolation. They do so for English, German, Latin, and both Norwegian datasets. It is also more often the case that the best standalone profile and contextualised approach yield the best ensemble when combined. Another notable difference is that, when used in isolation, profiles outperform XLM-R; the opposite is true in the ranking task.

Following the structure of Section~\ref{sec:results-ranking}, we first present the results of each approach individually and then we report the performance of ensembles.

\paragraph{Grammatical Profiles}
At least one of the three profile types is substantially above chance performance for each language; as in the ranking task, different profile types fit different datasets. Nevertheless, SYNT is the best profile type for 4 out of 7 datasets: German, Swedish, Norwegian-1, and Italian. For the first two, it achieves the best overall scores (see Table~\ref{tab:classification}). Combining morphology and syntax helps only in the case of Latin, where profiles obtain the best overall accuracy. 
%

\paragraph{Contextualised Embeddings}
The accuracy of APD and PRT is relatively similar across test sets, with the exception of Italian, where APD has the best overall accuracy and PRT is slightly below chance. Combining APD and PRT improves results for English, German, and Norwegian. The accuracy of clustering-based JSD is either close to or below chance level for all languages.

\paragraph{Ensembles}
Ensembles of grammatical profiles and contextualised embeddings are the best performing method for Norwegian. For German and Latin they are on par with pure profiles, and for English on par with pure XLM-R. The complementarity of different profile types and contextualised metrics varies across datasets yet it is overall stronger than that between profiles and static embeddings (combining the latter two improves performance only for Latin and for Norwegian-2, see Appendix~\ref{sec:app-static}). 
A more fine-grained analysis of the classification results of the ensembles reveals that 1) ensemble predictions are virtually always correct when the two standalone predictions also are, 2) ensembling tends to have positive effects on precision with respect to both standalone methods, and 3) it tends to improve the precision of contextualised methods.

\begin{table*}
    \centering
    \resizebox{\linewidth}{!}{
    \begin{tabular}{l|ccccccccc|c}
        \textbf{Method}	& \textbf{EN}	& \textbf{DE}	& \textbf{LA}	& \textbf{SW}	& \textbf{NO-1}	& \textbf{NO-2}	& \textbf{RU-1}	& \textbf{RU-2}	& \textbf{RU-3} & \textbf{AVG}\\
        \midrule
        \multicolumn{11}{c}{\textsc{PROFILES}} \\
        \midrule
        MORPH	    & 0.218	& 0.120	& 0.519	& 0.303	& 0.106	& \textit{0.409}	& 0.028	& \textit{0.241}	& \textit{0.293} & \textit{0.248} \\
        SYNT	    & \textit{0.331}	& 0.146	& 0.265	& 0.184	& \textit{0.179}	& 0.006	& \textit{0.056}	& 0.111	& 0.279 & 0.173 \\
        MORPHSYNT	& 0.320	& \textit{0.298}	& \textit{0.525}	& \textit{0.334}	& 0.064	& 0.265	& 0.000	& 0.149	& 0.242 & 0.244
        \\
       \midrule
        \multicolumn{11}{c}{\textsc{CONTEXTUALISED (XLM-R)}} \\
        \midrule 
        APD & \textit{\textbf{0.514}} & 0.073 & 0.162 & \textit{0.310} & 0.389 & \textit{0.387} & 0.372 & \textit{\textbf{0.480}} & \textit{\textbf{0.457}} & \textit{0.349} \\
        PRT & 0.320 & 0.210 & \textit{0.394} & 0.212 & 0.378 & 0.270 & 0.294 & 0.313 & 0.313 & 0.300 \\
        APD-PRT & 0.457 & 0.202 & 0.370 & 0.220 & \textit{\textbf{0.394}} & 0.325 & \textit{\textbf{0.376}} & 0.374 & 0.384 & \textbf{0.345} \\
        Clustering/JSD & 0.127  & \textit{0.287} & 0.318 &  -0.108& 0.160 & -0.137 & 0.247 & 0.267 &  0.362 & 0.169 \\
        \midrule
        \multicolumn{11}{c}{\textsc{ENSEMBLES}} \\
        \midrule
        APD-MORPH	          & 0.262 & 0.140 & 0.506 & 0.350 & 0.151 & \textit{\textbf{0.503}} & 0.062 & \textit{0.288} &  0.340 & 0.289\\
        APD-SYNT	          & 0.384 & 0.159 & 0.264 & 0.255 & \textit{0.262} & 0.119 & \textit{0.093} & 0.181 & \textit{0.354} & 0.230 \\
        APD-MORPHSYNT	      & \textit{0.390 }& \textit{0.290} & \textit{0.513} & \textit{\textbf{0.397}} & 0.180 & 0.364 & 0.036 & 0.216 & 0.299 & \textit{0.298}\\
        \hline
        PRT-MORPH	          & 0.278 & 0.204 & 0.528 & 0.305 & 0.236 & \textit{0.478} & 0.112 & \textit{0.309} & 0.336 & 0.309 \\
        PRT-SYNT             & 0.448 & 0.213 & 0.401 & 0.280 & \textit{0.351} & 0.146 & \textit{0.186} & 0.246 & \textit{0.351} & 0.291\\
        PRT-MORPHSYNT	      & \textit{0.451} & \textit{\textbf{0.354}} & \textit{\textbf{0.572}} & \textit{0.356} & 0.273 & 0.360 & 0.117 & 0.269 & 0.326 & \textit{0.342}\\
        \hline
        APD-PRT-MORPH    & 0.277 & 0.188 & 0.518 & 0.338 & 0.189 & \textit{0.497} & 0.092 & \textit{0.310} & 0.340 & 0.305\\
        APD-PRT-SYNT     & 0.405 & 0.189 & 0.376 & 0.295 & \textit{0.330} & 0.121 & \textit{0.147} & 0.235 & \textit{0.367} & 0.274\\
        APD-PRT-MORPHSYNT& \textit{0.418} & \textit{0.337} & \textit{0.554} & \textit{0.377} & 0.236 & 0.359 & 0.092 & 0.255 & 0.328  & \textit{0.328}\\        
    \end{tabular}}
    \caption{Spearman rank-correlation scores in the ranking task (`Task 2'). \textbf{Bold} indicates the best method overall (for each language); \textit{italic} indicates the best results for a group of methods.}
    \label{tab:ranking}
\end{table*}

\begin{table*}
    \centering
    \begin{tabular}{l|ccccccc|c}
        \textbf{Method}	& \textbf{EN}	& \textbf{DE}	& \textbf{LA}	& \textbf{SW}	& \textbf{NO-1}	& \textbf{NO-2}	& \textbf{IT} & \textbf{AVG} \\
         \midrule
        \multicolumn{9}{c}{\textsc{PROFILES}} \\
         \midrule
        MORPH	    & \textit{0.622}	& 0.479	& 0.625	& 0.581	& 0.486	& \textit{0.703}	& 0.500 & 0.571	\\
        SYNT	    & 0.514	& \textit{\textbf{0.625}}	& 0.514	& \textit{\textbf{0.677}}	& \textit{0.622}	& 0.514	& \textit{0.611} &	\textit{0.582} \\
        MORPHSYNT	& 0.541	& 0.521	& \textit{\textbf{0.675}}	& 0.581	& 0.486	& 0.432	& 0.444  & 0.526 \\
         \midrule
        \multicolumn{9}{c}{\textsc{CONTEXTUALISED (XLM-R)}} \\
         \midrule
        APD & 0.568 & 0.500 & 0.500 & \textit{0.613} & 0.486 & \textit{0.595} & \textit{\textbf{0.667}} & \textit{0.561} \\ 
        PRT & 0.595 & 0.500 & \textit{0.550} & 0.548 & 0.541 & 0.541 & 0.444 & 0.531 \\
        APD-PRT & \textit{\textbf{0.676}} & \textit{0.542} & \textit{0.550} & \textit{0.613} & \textit{0.568} & 0.459 & 0.500 & 0.558\\
        Clustering/JSD & 0.459 & 0.521 & 0.500 & 0.516 & 0.541 &  0.486 & 0.389 & 0.487 \\
         \midrule
        \multicolumn{9}{c}{\textsc{ENSEMBLES}} \\
         \midrule
        APD-MORPH	          & \textit{0.622} & 0.500 & 0.575 & \textit{0.613} & 0.541 & \textit{\textbf{0.730}} & 0.500 & 0.583 \\
        APD-SYNT	          & 0.568 & 0.479 & 0.550 & 0.581 & \textit{0.622} & 0.622 & \textit{0.611} & 0.576 \\   
        APD-MORPHSYNT	      & \textit{0.622} & \textit{\textbf{0.625}} & \textit{0.600} & \textit{0.613} & 0.514 & 0.703 & \textit{0.611} & \textit{\textbf{0.613}} \\ 
        \hline
        PRT-MORPH	          & \textit{\textbf{0.676}} & 0.458 & 0.525 & 0.581 & 0.541 & 0.486 & \textit{0.500} & 0.538 \\
        PRT-SYNT             & 0.541 & \textit{0.521} & \textit{0.575} & \textit{0.613} & \textit{\textbf{0.703}} & \textit{0.568} & \textit{0.500} & \textit{0.574} \\
        PRT-MORPHSYNT	      & 0.541 & 0.479 & 0.525 & 0.581 & 0.676 & 0.486 & 0.444 & 0.533 \\
        \hline
        APD-PRT-MORPH    & \textit{0.649} & 0.458 & \textit{0.650} & \textit{0.581} & 0.541 & \textit{0.676} & \textit{0.611} & \textit{0.595} \\ 
        APD-PRT-SYNT     & 0.514 & \textit{0.542} & 0.550 & 0.548 & \textit{0.676} & 0.595 & 0.500 & 0.561 \\
        PRT-MORPHSYNT	 & 0.541 & 0.479 & 0.525 & \textit{0.581} & \textit{0.676} & 0.486 & 0.444 & 0.533 \\
    \end{tabular}
    \caption{Binary accuracy scores in the classification task (`Task 1'). \textbf{Bold} indicates the best method overall (for each language); \textit{italic} indicates the best results for a group of methods.}
    \label{tab:classification}
\end{table*}

\subsection{Why ensembles fail}
\label{sec:results-why}
Tables~\ref{tab:ranking} and \ref{tab:classification} show that grammatical profiles are consistently worse than XLM-R on all Russian datasets, Norwegian-1 and English. This naturally extends to their ensembles, so for all these datasets, contextualised embeddings in isolation are the best approach. The explanation may seem simple for English: its poor morphology does not provide enough signal for semantic change detection.
Yet this does not hold for Russian (a synthetic language with rich morphology) and, arguably, for Norwegian. 
Moreover, ensembles with morphology-based grammatical profiles outperform pure XLM-R on Norwegian-2, but not on Norwegian-1. Thus, the explanation is likely not language-specific. 

We believe that the different nature of the diachronic corpora can be a better explaining factor. SemEval-2020 datasets feature time periods separated by at least several decades, and the same is true for Norwegian-2 (more than 20 years gap). In contrast, the gaps are much shorter for Norwegian-1 (5 years gap), Russian-1 and Russian-2 (2 years gap). We observe that when two time periods with a very short gap between them are compared, the distributions of morphosyntactic features largely overlap, negatively affecting 
the performance of grammatical profiles. In these cases, LM-based methods can still detect semantic change as they have access to lexical information: changes at the referential and topical level can happen much faster (consider, e.g., the words \textit{`computer'} or \textit{`mouse'} in English). On the other hand, when the gap between time periods is more substantial, changes in morphological and syntactic behavior of words also emerge. In these cases grammatical profiles help detect semantic shifts which LMs overlook. It is possible that adding the length of the time gap as a feature in our ensemble systems can make them less sensitive to the nature of the datasets .

Exceptions to this pattern are Latin (no gap between the time periods, but great performance of grammatical profiles) and Russian-3 (80 years gap, but profiles still lag behind XLM-R). For Latin, its extremely rich morphology can compensate for the small gap between time periods. Moreover, the second time period spans two millennia, making the short gap less problematic. Rich morphology does not help surpass XLM-R for Russian-3, but profiles do work much better for this dataset than for Russian-1 and Russian-2, where the gaps are only two years long.

\section{Analysis}
\label{sec:analysis}
In this section, we analyse the predictions of all methods beyond task performance. We quantitatively evaluate their complementarity  (Section~\ref{sec:analysis-correlation}), and investigate whether and how predictions made with grammatical profiles improve the performance of embedding-based metrics (Section~\ref{sec:analysis-qualitative}).

\subsection{Correlations between methods}
\label{sec:analysis-correlation}

\begin{figure}[t]
\includegraphics[width=1\columnwidth]{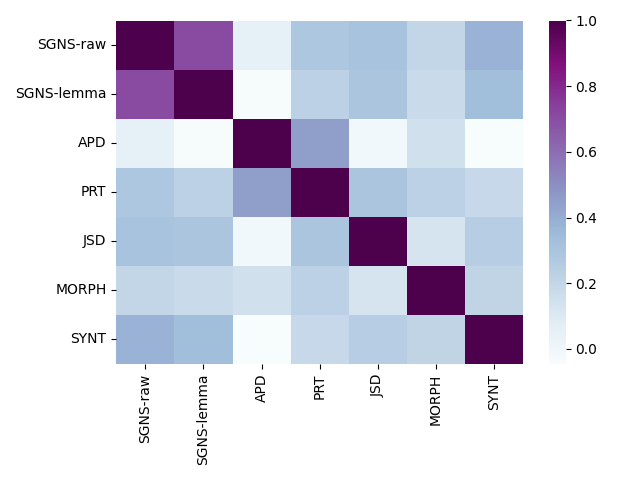}
\centering
\caption{Averaged Spearman correlations between model predictions.}
\label{fig:corr}
\end{figure}

To investigate whether various methods use different types of linguistic information, we compute Spearman rank-correlations between the predictions of standalone methods. The correlations, averaged over all datasets, are presented in Figure~\ref{fig:corr} (we show averaged correlations since they are highly consistent across corpora). We include the correlations of static embeddings as well (SGNS-raw and SGNS-lemma). More details about their implementations and performance can be found in Appendix~\ref{sec:app-static}.

The two methods with the highest correlation are SGNS-raw and SGNS-lemma. This is expected, as the two methods differ only in the lemmatisation of target words.
Profile-based methods (MORPH and SYNT) do not correlate with each other. Slight significant correlations are only observed for Russian-2 (0.32) and Russian-3 (0.45). Interestingly, for Russian, SYNT significantly correlates with static embeddings: the correlation with SGNS-raw is 0.48 for Russian-1, 0.52 for Russian-2 and 0.49 for Russian-3. Significant correlations between MORPH and static embeddings are observed for Latin (0.46) and Russian-3 (0.38). 

Contextualised methods correlate weakly with grammatical profiles. 
\hide{
JSD is significantly correlated with MORPH only for Russian-3 (0.33), PRT---with Russian-1 (0.27) and Russian-2 (0.29), while APD is not significantly correlated with MORPH for any of the datasets. SYNT is significantly correlated with JSD for Russian-1 (0.32) and Russian-2 (0.36), and significantly correlated with PRT for Russian-1 (0.26) and Russian-3 (0.35), though we did not observe any significant correlations between SYNT and APD for any of the datasets.
}
Although we once again observe exceptional behaviour for the Russian datasets, the correlation between profiles and contextualized embeddings is on average weaker than between profiles and static embeddings, which might explain why combining contextualized embeddings with profiles yields notable performance improvements.

\subsection{Qualitative analysis}
\label{sec:analysis-qualitative}
In this section, we inspect the error patterns of our methods to find out when grammatical profiles help correct the predictions of embedding-based metrics. We frame this analysis in terms of \textit{false positives} and \textit{false negatives}. The definition of false positives and negatives is straightforward in the classification task. For the ranking task, we look at the signed distance between gold and predicted rankings of each word, considering a word as a false positive when the positive distance is in the highest 20\% bin of the distance distribution (i.e., when the predicted rank is much higher than the true rank), and as a false negative if the negative distance is in the lowest 20\% bin (i.e., when the predicted rank is much lower than the true rank). For each language, we focus on the best grammatical profile, the best contextualised method, and the best ensemble of these two.

In the English ranking task, we observe that four of APD's five false positives are corrected by the ensemble: for example, the ranking of \word{tree} improves by 20 positions, that of \word{part} by 19, and that of \word{bag} by 17. As a result, \word{tree} and \word{bag} are only 1 position away from their respective gold ranks. For both words, the distribution of morphological features hardly vary between time period (e.g., 43.73\% of the usages of \word{tree} are singular, 56.27\% plural in the first time period; and in the second time period the percentages become 43.67\% and 56.33\%). Syntactic features vary only slightly; the most drastic change among these three words is the increase of direct object usages of \word{bag} from 33.16\% to 41.40\%, with all the other features remaining relatively stable---overall, a negligible change.
Among APD's five false negatives, four are corrected in the best ensemble (PRT-MORPHSYNT): the strongest ranking improvements concern \word{graft} and \word{plane}, whose rankings improve respectively by 18 and 15 positions. The syntactic profiles of these words vary substantially across time periods, with multiple syntactic categories increasing or decreasing their frequency of usage (e.g., usages of \word{plane} in subject and object position increase from 12.85\% to 24.13\% and from 13.25\% to 19.67\% respectively; while usages as a noun modifier decrease from 35.34\% to 20.36\%).
The two targets that do not benefit from the ensemble are \word{gas} and \word{risk}: both are false negatives for the best grammatical profile (SYNT) and they remain for the ensemble.

In the Norwegian ranking task, the best ensemble (APD-MORPH) helps pure APD mostly by fixing extreme false positives and false negatives. As an example of a fixed false positive, APD ranked \word{test}  (\sense{test}) very high, although in fact it did not experience any semantic change at all (change score of 0). APD-MORPH decreased the change score assigned to \word{test} from 0.216 down to 0.013, returning it to its proper place at the bottom of the ranking. On the other hand, \word{stryk} changed its dominant meaning sharply in the 21st century from \sense{river rapids} to \sense{failure}, but APD failed to capture it. APD-MORPH fixed this false negative by moving  \word{stryk} significantly upwards in the ranking, only 8 position away from its gold rank.

In the Latin predicted rankings, it is somewhat likely for a word to be a false positive --- e.g., \word{itero} (\sense{to repeat}), \word{jus} (a \sense{right}, the \sense{law}) \word{ancilla} (\sense{handmaid}) --- or a false negative --- \word{virtus} (\sense{strength}; \sense{courage}; \sense{manliness}), \word{humanitas} (\sense{human nature}; \sense{kindness}; \sense{civisilation}), \word{pontifex} (\sense{bishop}; but also, the \sense{POPE}) --- for both contextualised embeddings (PRT) and profiles (MORPHSYNT). As in the case of English, the ranking of these words does not improve with ensembling (PRT-MORPHSYNT). Overall, 7 out of 10 false negatives and 2 out of 7 false positives are corrected by the ensemble.

For German, too, ensembling (PRT-MORPHSYNT) is most effective for false negatives, 5 out of 8 are corrected. The words with the greatest improvements are \word{abdecken} (\sense{to uncover}; but also, in financial jargon, \sense{to cover}, as in \textit{Risiko abdecken}, \textit{to cover a risk}), gaining 16 positions, and \word{Eintagsfliege} (\sense{mayfly}, the insect; but also, metaphorically, \sense{fleeting star}) gaining 17 positions and thereby obtaining the exact gold rank. Nevertheless, out of 8 false positives, 4 are corrected; with, e.g., \word{aufrechterhalten} (\sense{to sustain}) losing 31 rankings and \word{Festspiel} (\sense{festival}) losing 18. As we observed for English and Latin, some words are simply difficult to rank for both methods (here, JSD and PRT-MORPHSYNT): for example, the degree of semantic change of \word{Truppenteil} (\sense{troop unit}) and \word{Lyzeum} (\sense{lyceum}) is overestimated whereas the change of \word{packen} (\sense{to pack}; \textit{to seize}) and \word{vorliegen} (\sense{to be available}; \sense{to be existent});  is underestimated.

For Italian, we analyse the classification task. Among APD's 4 false positives, 2 are corrected in the ensemble's ranking (APD-SYNT), \word{processare} (\sense{process}; \sense{take to trial}) and \word{unico} (\sense{unique}); two nouns, \word{brama} (\sense{yearning}) and \word{cappuccio} (\sense{hood}), remain misclassified. APD's false negatives are \word{pilotato} (\sense{driven}; but also, metaphorically, \sense{premeditated}) and \word{rampante} (\sense{unbridled}; metaphorically, \sense{exuberant}); \word{pilotato} is correctly classified by the ensemble while \word{rampante} remains undetected by all methods. Overall, the contribution of \textit{SYNT} is not always helpful: it also leads to one changing word being labelled as stable, and three stable words being classified as changing.

\section{Conclusion}
\label{sec:conclusion}

We showed that providing large pre-trained language models with explicit morphosyntactic information can in many cases help detect and quantify lexical semantic change. Such `ensemble' predictions are produced in a very straightforward way---i.e., by computing the geometric mean between semantic change scores predicted by grammatical profiles and by language models (via their contextualized embeddings). In the majority of the datasets under analysis (treating the three Russian datasets as one), the ensemble predictions outperformed single grammatical profiles or contextualised embeddings in the task of ranking words by the degree of their semantic change. The datasets where this was not true are characterized by specific properties: either languages with poor morphology or long time spans separated by narrow gaps. 

We believe this means that although Transformer-based language models (like XLM-R, which we used here) are able to track morphological and syntactic properties to some extent~\cite{warstadt-etal-2020-blimp-benchmark}, their encoding of grammatical features is only approximate and can therefore be improved by explicit linguistic pre-processing (morphological tagging and syntactic parsing). At any rate, we showed that this is true for the semantic change detection task, when a model has to take into account diachronic changes in morphosyntactic properties of words. The signal provided by these changes is complementary to the changes in typical lexical contexts more easily captured by distributional language models. Thus, it is still too early to fire the linguist, even if the `linguist' is in fact an automated tagger.

As has already been said, an important limitation of grammatical profiles is their low performance when measuring semantic change across long time periods separated by very narrow gaps. This makes sense from a linguistic point of view: grammar changes slowly and gradually, sharp bursts are rare. In contrast, lexical contexts can change very quickly: for example, due to social and political events or technical progress, which is why language models excel with these datasets. The main practical take-away is therefore that diachronic grammatical profiles should be used in combination with language models especially when the gap between the compared time periods is large enough for significant grammatical changes to occur. 

In the future, we plan to experiment with more sophisticated ensembling methods that go beyond simple averaging (including the usage of the information about gaps between time spans), and to perform a deeper analysis of ensemble predictions, especially in relation to distinct word senses. Finally, we also plan to evaluate ensembles formed with monolingual language models, instead of the multilingual XLM-R, as they have the potential to better capture the idiosyncrasies of specific languages.

\section*{Acknowledgements}

This work has been partly supported by the European Union’s Horizon 2020 research and innovation programme under grant 819455 (DREAM) and by Academy of Finland under grant numbers 333716 and 333717 (RiCEP). Our experiments were run on resources provided
by UNINETT Sigma2 - the National Infrastructure for High Performance Computing and Data Storage in Norway.

\bibliography{anthology,custom}
\bibliographystyle{acl_natbib}

\newpage
\appendix

\section*{Appendix}
\label{sec:appendix}

\section{Contextualised embeddings}
\label{sec:app-contextualised}
Given two time periods $t_1, t_2$ , two corpora $C_1, C_2$, and a set of target words, we use a neural language model to obtain \textit{token embeddings} of each occurrence of the target words in $C_1$ and $C_2$ and use them to compute a continuous change score. This score indicates the degree of semantic change undergone by a word between $t_1$ and $t_2$. As a language model, we choose XLM-R~\cite{conneau-etal-2020-unsupervised}, pre-trained multilingual transformer, in the Huggingface implementation~\cite{wolf-etal-2020-transformers}. 

\subsection{Target Lemmas and Word Forms}
\label{sec:app-targets}
The lists of target words that we rely on contain annotations for lemmas. However, only extracting embeddings for exact matches of the lemmas would result in discarding a large number of word usages, those where the target lemma takes another form (e.g., as a result of grammatical inflection). To take all of a lemma's possible word forms into account, we  parse the corpora using UDPipe~\cite{straka-strakova-2017-tokenizing} and collect a set of word forms for each target word from the UDPipe output. 
Furthermore, because some word forms are not present in the vocabulary of XLM-R, we add them to the vocabulary before fine-tuning.\footnote{Even after adding the word forms to the vocabulary, the Huggingface tokenizer still fails to recognise about a dozen of the target word forms and splits them into sub-tokens. For these exceptional cases, we extract the average contextualised embedding over the sub-tokens.}

\subsection{Finetuning the Language Model}
\label{sec:app-finetuning}
As a first step, to adapt the model to the characteristics of the diachronic corpora, we finetune it, separately, on each language-specific corpus. We limit the maximum sequence length of the transformer to 256 and train the model with a batch size of 16 for an amount of epochs dependent on the corpus size: 5 epochs for English and Latin, 3 for German and Swedish, 2 for Russian, Italian and Norwegian. 

\subsection{Extracting contextualised embeddings}
\label{sec:app-extraction}
Given a target word $w$ and its sentential context $s = (v_1, ..., v_i, ..., v_m)$ with $w = v_i$, we extract the activations of the language model's hidden layers for sentence position $i$. We then average over the layers (12 for XLM-R) and obtain a single vectorial representation (for XLM-R, the vector dimensionality is 768). In our experiments, the maximum context length $m$ is set to 256 and sentences are processed in batches of size 32. The $N_w$ contextualised embeddings collected for $w$ can be represented as the usage matrix $\textbf{U}_w = \left( \textbf{w}_1, \ldots, \textbf{w}_{N_w} \right)$. The time-specific usage matrices $\textbf{U}_w^1, \textbf{U}_w^2$ for time periods $t_1$ and $t_2$ are used as input to a metric of semantic change. 

\subsection{Metrics of Semantic Change}
\label{sec:app-metrics}
As explained in Section~\ref{sec:method-contextualised}, semantic change scores are computed using three metrics: 1) average pairwise distance or APD~\cite{giulianelli-etal-2020-analysing}, 2) prototype distance or PRT~\cite{kutuzov-giulianelli-2020-uio}, and 3) Jensen-Shannon Divergence between embedding cluster distributions or JSD~\cite{martinc2020capturing,giulianelli-etal-2020-analysing}:

\paragraph{APD}
Given two usage matrices $\textbf{U}_w^{t_1}, \textbf{U}_w^{t_2}$, the degree of change of $w$ is calculated as the average cosine distance between any two embeddings from different time periods:
\begin{align}
\resizebox{\linewidth}{!}{$
    \operatorname{APD}\left(\textbf{U}_w^{t_1}, \textbf{U}_w^{t_2}\right) =  \frac{1}{N_w^{t_1} \cdot N_w^{t_2}} \sum_{\textbf{x}_i \in \textbf{U}_w^{t_1},\ \textbf{x}_j \in \textbf{U}_w^{t_2}} cos\left(\textbf{x}_i, \textbf{x}_j\right)
$}
\end{align}
where $N_w^{t_1}$ and $N_w^{t_2}$ are the number of occurrences of $w$ in time periods $t_1$ and $t_2$. 

\paragraph{PRT}
Here, the degree of change of $w$ is measured as the cosine distance between the average token embeddings (`prototypes') of all occurrences of $w$ in the two time periods:
\begin{align}
\resizebox{\linewidth}{!}{$
    \operatorname{PRT}\left(\textbf{U}_w^{t_1}, \textbf{U}_w^{t_2}\right) = 1 - cos\left(\frac{\sum_{\textbf{x}_i \in \textbf{U}_w^{t_1}} \textbf{x}_i}{N_w^{t_1}}, \frac{\sum_{\textbf{x}_j \in \textbf{U}_w^{t_2}} \textbf{x}_j}{N_w^{t_2}}\right) 
$}
\end{align}

\paragraph{JSD}
To compute this measure, we form a single usage matrix $[\textbf{U}^{t_1}_w ; \textbf{U}^{t_2}_w]$ with occurrences from two corpora. We standardise it and then clustered its entries using Affinity Propagation~\cite{frey2007affinity}, a clustering algorithm which automatically selects a number of clusters for each word.\footnote{We use the scikit-learn implementation of Affinity Propagation with default hyperparameters.}
Finally, we define probability distributions $\textbf{u}_w^{t_1}, \textbf{u}_w^{t_2}$ based on the normalised counts of word embeddings in each cluster and compute a the Jensen-Shannon Divergence~\cite{lin1991jsd} between the distributions:
\begin{align}
\resizebox{\linewidth}{!}{$
\operatorname{JSD}(\textbf{u}_w^{t_1}, \textbf{u}_w^{t_2}) = \operatorname{H}\left(\frac{1}{2} \left( \textbf{u}_w^{t_1} + \textbf{u}_w^{t_2}\right) \right) - \frac{1}{2} \left(\operatorname{H}\left(\textbf{u}_w^{t_1}\right) - \operatorname{H}\left(\textbf{u}_w^{t_2}\right)\right)
$}
\end{align}

\section{Static Embeddings}
\label{sec:app-static}
We follow the common approach proposed by~\citet{hamilton-etal-2016-diachronic}, SGNS+OP, to train skip-gram negative sampling embeddings \cite[SGNS;][]{Mikolov_representation:2013} from scratch for each time period of the diachronic corpus, and then to align the separate vector spaces using the Orthogonal Procrustes method (OP). Semantic change is measured as the cosine distance between the embeddings of a target word in the aligned spaces~\cite[for more details, see][]{schlechtweg-etal-2019-wind}.

We decided not to lemmatize our corpora for these experiments to preserve as much grammatical information as it is possible but we use two preprocessing strategies for target words. In the first strategy (\textbf{SGNS-raw}) we use a raw, unlemmatized, corpus and learn embeddings for target words only in their dictionary form. All other inflected forms of the target words are ignored. In the second strategy (\textbf{SGNS-lemma}), we lemmatize target word occurrences (but not other words) and thus use all target word forms to train their embeddings.

In the ranking task, SGNS+OP confirms itself as a very competitive approach, achieving the best correlation scores on German, Swedish, and Russian 3 (see Table~\ref{tab:ranking_sgns}). Our results show that lemmatizing target word forms, so that they all contribute to the same static embedding, brings substantial performance improvements as well as more stability across test sets.

\begin{table*}
    \centering
    \resizebox{\linewidth}{!}{
    \begin{tabular}{l|ccccccccc}
        \textbf{Method}	& \textbf{EN}	& \textbf{DE}	& \textbf{LA}	& \textbf{SW}	& \textbf{NO-1}	& \textbf{NO-2}	& \textbf{RU-1}	& \textbf{RU-2}	& \textbf{RU-3} \\
     \midrule
        \multicolumn{10}{c}{\textsc{STATIC}} \\
        \midrule 
        Raw text (SNGS-raw)	& 0.378 & 0.226 & 0.250 & -0.036 & 0.320 & 0.181 & 0.101 & 0.148 & 0.255 \\
        Target words lemmatized (SGNS-lemma) & 0.498 & \textbf{0.369} & 0.106 & \textbf{0.494}	& 0.238 & 0.392 & 0.256	& 0.292	& \textbf{0.538} \\
                \midrule
        \multicolumn{10}{c}{\textsc{ENSEMBLES}} \\
        \midrule
        SGNS-raw-MORPH & 0.253 & 0.105 & 0.436 & 0.204 & 0.116 & 0.368 & 0.020 & 0.222  & 0.275 \\
        SGNS-raw-SYNT & 0.341 & 0.159 & 0.234 & 0.158 & 0.250 &  0.024 & 0.019 & 0.113 & 0.248  \\
        SGNS-raw-MORPHSYNT & 0.354 & 0.258 & 0.454 & 0.297 & 0.142 & 0.218 & 0.013 & 0.148 & 0.229  \\
        SGNS-lemma-MORPH & 0.255 & 0.157 & 0.409 & 0.386 & 0.106 & 0.440 & 0.057 & 0.259 & 0.332 \\
        SGNS-lemma-SYNT & 0.364 & 0.173 & 0.224 & 0.242 & 0.212 & 0.156 & 0.071 & 0.129 &  0.315 \\
        SGNS-lemma-MORPHSYNT & 0.367 & 0.269 & 0.415 & 0.461 & 0.128 & 0.341 & 0.023 & 0.163 & 0.286  \\
    \end{tabular}}
    \caption{Spearman correlation scores in the ranking task (`Task 2') with type-based static embeddings (SGNS-OP). Bold values are cases when SGNS-OP outperforms all other methods (XLM-R and grammatical profiles).}
    \label{tab:ranking_sgns}
\end{table*}

Our classification results again confirm the strength of static embeddings, which outperform other approaches for German, Norwegian-1, and Italian (for English and Swedish, they perform on par with the profile-contextualised ensembles). Target word form lemmatization is important but less decisive than in the ranking task (see Table~\ref{tab:classification_sgns}).


\begin{table*}
    \centering
    \begin{tabular}{l|ccccccc}
        \textbf{Method}	& \textbf{EN}	& \textbf{DE}	& \textbf{LA}	& \textbf{SW}	& \textbf{NO-1}	& \textbf{NO-2}	& \textbf{IT} \\
         \midrule
        Raw (SGNS-raw)	& 0.514 & 0.542 & 0.400 & 0.548	& \textbf{0.757} & 0.649 & 0.722 \\
        Target words lemmatized (SGNS-lemma) & \textbf{0.676} & \textbf{0.646} & 0.375 & \textbf{0.742}	& 0.676 & 0.676 & \textbf{0.778} \\
         \midrule
        \multicolumn{8}{c}{\textsc{ENSEMBLES}} \\
         \midrule
        SGNS-raw-MORPH & 0.622 & 0.562 & 0.600 & 0.484 & 0.486 & 0.622 & 0.500 \\
        SGNS-raw-SYNT & 0.541 & 0.583 & 0.550 & 0.581	& 0.649 & 0.486 & 0.500 \\
        SGNS-raw-MORPHSYNT & 0.649 & 0.625 & 0.500 & 0.677 & 0.595 & 0.486 & 0.611 \\
        SGNS-lemma-MORPH	& 0.622 & 0.438 & 0.625 & 0.484	& 0.514 & 0.703 & 0.500 \\
        SGNS-lemma-SYNT & 0.541 & 0.479 & 0.525 & 0.581 & 0.649 & 0.568 & 0.611 \\
        SGNS-lemma-MORPHSYNT & 0.649 & 0.604 & 0.600 & \textbf{0.742} & 0.514 & 0.676 & 0.389 \\
    \end{tabular}
    \caption{Binary accuracy scores in the classification task (`Task 1') with type-based static embeddings (SGNS-OP). Bold values are cases when SGNS-OP outperforms all other methods (XLM-R and grammatical profiles).}
    \label{tab:classification_sgns}
\end{table*}

\end{document}